\documentclass[10pt,twocolumn,letterpaper]{article}

\usepackage[pagenumbers]{cvpr} 

\usepackage{todonotes}

\definecolor{cvprblue}{rgb}{0.21,0.49,0.74}
\usepackage[pagebackref,breaklinks,colorlinks,allcolors=cvprblue]{hyperref}
\usepackage{multirow}

\def\paperID{10998} 
\def\confName{CVPR}
\def\confYear{2025}

\title{PVCap: Towards Accurate 3D Dense Captioning via PseudoCap and VoxelCapNet}

\author{
  Xiaopei Wu\textsuperscript{1,2}, Chenshu Hou\textsuperscript{2}, Liang Peng\textsuperscript{2}, Dan Xu\textsuperscript{3}, Binbin Lin\textsuperscript{2}, Xiaoshui Huang\textsuperscript{4} \\ 
  Yuenan Hou\textsuperscript{1}, Yu Li\textsuperscript{1}, Wenxiao Wang\textsuperscript{2}, Haifeng Liu\textsuperscript{2}, Deng Cai\textsuperscript{2}, Wanli Ouyang\textsuperscript{1}
  \vspace{0.3em} \\
  \textsuperscript{1}SH AI Lab \textsuperscript{2}ZJU \textsuperscript{3}HKUST \textsuperscript{4}SJTU
}

\begin{document}
\maketitle
\def\algorithmname{MethodName}

\begin{abstract}
3D dense captioning, an emerging vision-language task, aims to generate descriptive sentences 
for each object in the 3D scene. Despite the impressive results achieved by previous methods, 
they suffer from two limitations.
First, current research often employs global rigid transformations, such as rotation, to augment scenes without changing their spatial layouts.
However, diverse spatial layouts are crucial for training a 3D dense captioning model to describe spatial relations between objects.
Second, previous works mainly focus on the design of the caption generation pipeline while 
utilizing a simple network architecture for other components, \textit{i.e.}, backbone and detection head, 
which is crucial for extracting rich semantic information for captioning.
In this paper, we propose \textbf{PVCap} to alleviate the aforementioned problems. 
Our PVCap consists of PseudoCap and VoxelCapNet.
Specifically, PseudoCap employs a random mixing technique on instances within the dataset, 
generating numerous pseudo frames with diverse spatial layouts at the instance level. 
By utilizing a teacher-student framework, PseudoCap obtains pseudo caption labels for these pseudo frames. 
This data augmentation approach significantly increases the number of training samples and 
enhances the model's ability to describe the environment effectively.
Regarding VoxelCapNet, we introduce a robust caption network that utilizes voxel features and 
adapts the caption head to the voxel-based network architecture. Our VoxelCapNet can serve as 
a competitive baseline for future research on 3D dense captioning.
Extensive experiments are conducted on two prevalent benchmarks, \emph{i.e.}, ScanRefer and Nr3D. 
Notably, our method surpasses current state-of-the-art by \textbf{11.41\%} and \textbf{13.99\%} 
in CIDEr@0.5IoU, respectively. 
Codes will be made publicly available.
\vspace{-2mm}

\end{abstract}

\section{Introduction}
\label{sec:introduction}
3D dense captioning \cite{scan2cap,d3net,x-trans2cap,spacap3d,more,3djcg} aims to describe 
individual objects in 3D point cloud scenes by natural language. 
It can be divided into two tasks: object detection and object caption generation.
Scan2Cap\cite{scan2cap}, MORE\cite{more} and SpaCap3D\cite{spacap3d} propose well-designed 
relation reasoning modules to model relations among object proposals efficiently.
\cite{contextual} introduces contextual information from two branches to improve the caption.
3DJCG\cite{3djcg} and D3Net\cite{d3net} study the correlation between 3D visual grounding 
and 3D dense captioning and demonstrate the synergistic effect of these two tasks.
Additionally, 
Vote2Cap\cite{vote2cap} and Vote2Cap++\cite{vote2cap++} 
jointly train object detection and caption generation, enabling the mutual promotion of the two tasks.

Though previous methods have achieved remarkable results,
they suffer from two primary issues: insufficient data augmentation and poor network architecture.
Regarding data augmentation, current 3D dense captioning methods often rely on 
rigid transformations (such as rotation). 
However, these transformations do not alter the spatial layout of the scene, and 
describing the environment surrounding an object is crucial for accurate captioning. 
Simple data augmentation techniques provide only limited variation in the data space available for the caption model, which restricts its ability to generate precise captions.
As to the network architecture, we observe that existing works primarily focus on the 
design of the caption generation pipeline while ignoring the importance of other components.
They usually leverage a simple network architecture, such as PointNet++\cite{pointnet++} 
as the backbone and 3DETR\cite{3detr} as the detection head, to extract features for the caption head.
These architectures cannot provide sufficient information for accurate captioning, 
leading to unsatisfactory results.

\begin{figure}[t]
\centering
\includegraphics[width=0.41 \textwidth]{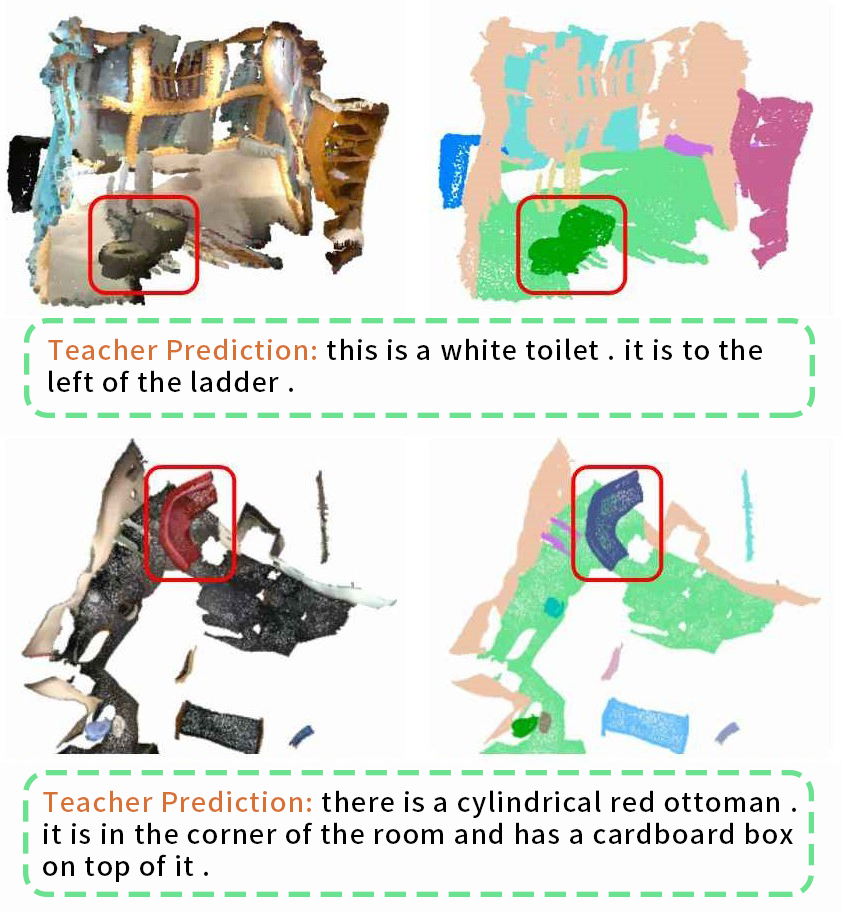} 
\vspace{-4pt}
\caption{The visualization of pseudo frames and their predicted captions generated by the proposed PseudoCap.}
\label{fig:pseudo}
\vspace{-12pt}
\end{figure}

To alleviate these issues, we propose a novel 3D dense captioning framework dubbed PVCap. 
For the data augmentation issue, we present a strong data augmentation strategy tailored for 3D dense captioning, 
termed PseudoCap. It creates an instance database by cropping instances from the dataset.
During the training stage, instances are randomly selected from the database and shuffled to generate pseudo frames, as shown in Figure \ref{fig:pseudo}.
These pseudo frames exhibit diverse spatial layouts, providing sufficient supervision for the caption model 
to improve its ability to describe the environment. 
As PseudoCap shuffles objects, making original captions invalid for the pseudo frame, we apply a teacher-student framework for proper caption supervision.
The teacher is a frozen caption model pre-trained on the 
original caption dataset. We first feed the pseudo frames to the teacher model 
to generate pseudo caption labels, then filter high-quality pseudo caption labels 
as the supervision of pseudo frames to 
train the student. Our PseudoCap significantly augments the caption dataset and can reduce 
the expensive cost associated with dense caption labeling.

In terms of network architecture, current architectures for point cloud understanding can be 
roughly categorized into voxel-based and point-based methods. 
 A popular viewpoint is that the point-based methods can yield higher performance and the voxel-based ones 
are more efficient. However, recent successes in 3D segmentation \cite{minkunet,swin3d,ptv2} and 3D object detection \cite{fcaf3d,cagroup3d} demonstrate that voxel-based methods are both efficient and effective.
To this end, we present the 
first voxel-based 3D dense caption network, named VoxelCapNet. VoxelCapNet is simple but effective.
The powerful voxel-based backbone extracts voxel features, 
while the detection head filters object features based on confidences and applies non-maximum suppression (NMS). These features are then 
fed into the caption head to generate captions. 
Experiment results show that our design significantly 
improves the caption performance. We believe that our VoxelCapNet can serve as a strong baseline 
to advance the field of 3D dense captioning.

Experiments on two popular datasets, \emph{i.e.}, ScanRefer and Nr3D, shows 
that our approach surpasses previous state-of-the-art by a large margin, which strongly 
demonstrates the superiority of our method. To summarize, the main contributions of this work are listed as follows:
\begin{itemize} 
\setlength\itemsep{0.5em}
    \item We propose a novel data augmentation, called PseudoCap, for 3D dense captioning, which provides 
    diverse spatial layouts to benefit the learning of the 3D dense captioning model on spatial relations description.

    \item We present a novel voxel-based caption network, VoxelCapNet, which adapts the caption model
    to voxel-based network architectures and shows impressive results. It can serve as a strong baseline for 3D dense captioning.

    \item Extensive experiments show that our proposed PVCap delivers the new state-of-the-art performance on 
    both ScanRefer (89.57\% C@0.5IoU) and Nr3D (61.61\% C@0.5IoU) benchmarks.
	
\end{itemize} 



\section{Related Work}
\label{sec:relatedwork}

\noindent \textbf{3D Dense Captioning.}
3D Dense Captioning (3DDC) is an emerging task that requires a model to accurately 
localize all objects in a complex 3D scene using bounding boxes and generate descriptive 
captions for them. This task is particularly challenging due to the orderless, sparse 
and irregular characteristics of 3D point clouds. 
\cite{scan2cap, d3net, reman, more}
regard every box estimation from proposed 3D detectors 
as a graph node, constructing $K$ nearest neighbor graphs manually to extract features through 
graph operations. 
\cite{spacap3d} replace graph operations with a spatiality-guided transformer to capture spatial relations. 
\cite{3d-vlp, unit3d, 3d-vista, 3djcg} 
explore the mutual enhancement between 3D vision-language tasks and fundamental 3D vision tasks like object detection and semantic segmentation.
Vote2Cap-DETR \cite{vote2cap} 
introduces an end-to-end 3D dense captioning method for the first time and abandons the previous 
"detect-then-describe" pipeline. Vote2Cap-DETR++~\cite{vote2cap++} further decouples the queries 
into localization and caption queries to generate different understandings of the 3D environment from separate perspectives.
Although previous approaches have shown impressive results, their limitations arise 
from \emph{insufficient data augmentation} and \emph{poor network architecture}.
Our proposed PVCap stands apart from existing works by introducing a novel data augmentation 
method to generate pseudo frames with diverse spatial layouts, along with a voxel-based network 
to supply rich features for accurate captioning.

\noindent \textbf{3D Object Detection.}
Existing 3D object detection approaches could be divided into two categories regarding 
point cloud representations, \textit{i.e.}, voxel-based methods and point-based methods.
Popularized by PointNet \cite{pointnet}, point-based methods \cite{votenet,brnet,h3dnet,rbgnet} 
are widely used in estimating bounding boxes directly from raw point clouds.
Although these methods allow for direct feature extraction by analyzing point neighbors 
and require relatively low memory consumption, their application to large-scale point clouds 
is challenging due to the density variations. Voxel-based methods \cite{voxelnet,second,fcaf3d,cagroup3d} 
transform point clouds to regular 2D or 3D girds, then traditional 2D convolution and 
efficient 3D sparse convolution can be used to extract grid features.  
The recent success of voxel-based methods on 3D segmentation \cite{minkunet,swin3d,ptv2,octnet,moe3d,sega3d,taseg,pvkd} 
and 3D object detection \cite{fcaf3d,cagroup3d,nerf-det++} demonstrates that voxel-based methods have great 
potential in terms of both efficiency and accuracy. Inspired by these works, we leverage the 
voxel-based detection head to enhance our network architecture for 3D dense captioning.

\noindent \textbf{2D Vision-Language tasks.}
Image captioning requires a model to generate textual descriptions for images. 
Previous approaches can be roughly divided into two lines, i.e., trained-from-scratch methods 
and pre-training methods. For the trained-from-scratch methods 
\cite{dou2022empirical,jiang2018recurrent,li2022blip,sad,yang2022vision}, 
researchers mainly focus on enhancing model architecture and/or refining training losses, 
with a specific emphasis on addressing the image captioning task.
For the pre-training methods \cite{chen2020uniter,hu2022scaling,li2021align,li2020oscar,su2019vl}, 
the primary emphasis lies in how to effectively pre-train the model on a large image-text corpus 
so that it can transfer well to a broad set of downstream vision-language tasks.
\cite{jain2021perturb} employs semi-supervised learning (SSL) 
methods with noisy student training, conducting effective feature augmentation techniques. \cite{kuang2023dlip} trains a fused vision-language model 
with multi-modal distillation to develop a light, efficient model.
Though these methods are effective in captioning 2D scene images, 
they cannot be directly applied to 3D dense captioning because they fail to fully leverage 
sufficient 3D information for localizing and describing 3D objects.
In this paper, we tailor an effective data augmentation method and a strong voxel-based network for 3D dense captioning, achieving impressive results.

\section{Methodology}
\label{sec:methodology}
\begin{figure*}[t]
\vspace{-4mm}
	\begin{center}
		\setlength{\fboxrule}{0pt}
		\fbox{\includegraphics[width=.99\textwidth]{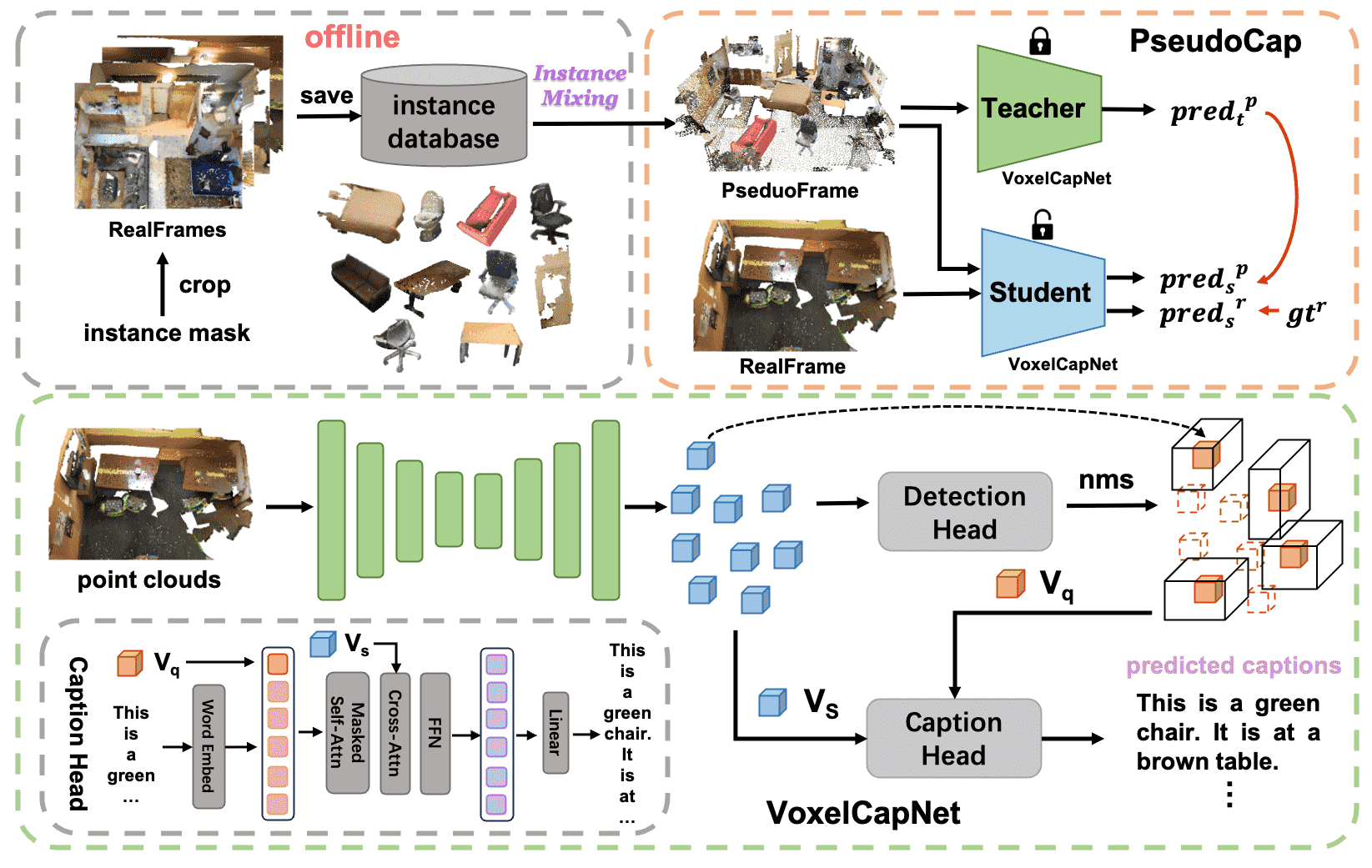}}
	\end{center}
	\vspace{-6mm}
	\caption{\textbf{Framework overview of our PVCap.} 
 Our PVCap consists of two key components: (1) \textbf{PseudoCap} mixes instances from different scenes to generate pseudo frames with diverse spatial layouts, providing extensive training data to enhance the ability of the model on spatial relation captioning. 
 To prepare caption supervision for pseudo frames, we leverage a frozen teacher to generate 
 pseudo caption labels. Finally, the student is trained with both pseudo frames and real frames. 
 (2) \textbf{VoxelCapNet} improves the caption model from the perspective of the network architecture. 
 We make full use of the strong representation ability of the voxel-based backbone and detection head 
 to supply rich features for the caption head. The caption head follows the design of \cite{vote2cap}.} 
	\label{fig:overview}
	\vspace{-4mm}
\end{figure*}

\subsection{Problem Definition and Framework Overview}
The objective of 3D dense captioning is to localize objects of interest accurately 
and generate natural language descriptions for each object in a given 3D scene. 
In this task, our model takes an input point cloud denoted as 
$P = \left[p_{in}; f_{in}\right] \in \mathbb{R}^{N \times \left(3 + D\right)}$, 
which represents an indoor 3D scene. Here, $N$ is the number of points, 
$p_{in} \in \mathbb{R}^{N\times 3}$ is the 3D coordinates of each point, 
and $f_{in} \in \mathbb{R}^{N\times D}$ is additional feature for each point, 
such as color, normal or multiview features introduced by \cite{scan2cap, scanrefer}. 
The output is a set of box-caption pairs $( B, C ) = \{(b_1, c_1), \cdots, (b_K, c_K)\}$, 
which represents an estimation of $K$ distinctive objects.

As illustrated in Figure~\ref{fig:overview}, our PVCap consists of two key components:
PseudoCap and VoxelCapNet. VoxelCapNet builds the caption head on voxel features 
from the voxel-based backbone and detection head, as shown at the bottom 
of Figure~\ref{fig:overview}. PseudoCap further enhances the model's captioning ability  
by leveraging large amounts of pseudo frames with diverse spatial layouts. 
Note that both the teacher and student in PseudoCap utilize the VoxelCapNet.

\subsection{PseudoCap}
\label{sec:pseudo-caption}
Data augmentation is an important technique to relieve data scarcity and improve model performance.
Yet, existing data augmentations for 3D dense captioning cannot provide diverse spatial layouts, limiting the ability of the model to describe environments effectively. 
To this end, we tailor an effective data augmentation strategy for 3D dense captioning, termed PseudoCap.
Here, we provide detailed introduction.

\subsubsection{Instance Database.}
To generate various spatial layouts, we first need to manipulate all instances in the dataset.
To this end, we build an instance database to store point clouds of all instances.
It uses ground truth bounding boxes or instance masks to crop point clouds of each instance 
in the dataset, and the cropped instances will be collected to generate 
a gt database $\mathcal{D} = \{s_i\}_{i=1}^{N_{gt}}$. Here, $N_{gt}$ is the number of instances 
in the whole dataset, and $s_i$ denotes the $i^{th}$ instances. 
For $s_i$, it consists of the corresponding point clouds ${p}_i$ and 
its labels ${l}_i$, \textit{i.e.} semantic category and bounding box.

\subsubsection{Instance Mixing.}
Given the instance database $\mathcal{D}$, we can freely assemble different instances to 
generate pseudo frames with diverse layouts. Specifically, a pseudo frame $\mathcal{P}$ 
starts from a real frame $\mathcal{R}$ of the dataset.
$\mathcal{P}$ inherits the background point clouds of $\mathcal{R}$ but filters the 
foreground point clouds of $\mathcal{R}$. This is necessary because existing instances 
of $\mathcal{R}$ may exclude new instances from joining the scene due to collision detection. 
Then, we randomly select several instances from the instance database $\mathcal{D}$ for each category 
and paste them to $\mathcal{P}$. To prevent pasted instances from overlapping with existing instances 
in the scene, we need to perform collision detection. Specifically, if a new instance 
has an IoU larger than a pre-defined threshold with that of any existing instance, it will not be pasted 
to the scene. Besides, with more instances joining, the scene will become crowded. 
If we choose instances according to a fixed category queue when we perform Instance Mixing, 
there may be no space for categories at the end of the queue, which will cause inter-class imbalance. 
To solve the issue, we simply shuffle the category queue every time we use Instance Mixing.

\subsubsection{Teacher-Student Framework.}
When an instance is pasted to a new environment, its detection label can be reused while 
its caption label becomes invalid because the caption label is related to the surroundings of an instance.
To address the caption label invalidation issue of pseudo frames, we leverage a teacher-student framework.
The teacher and student use the same caption model and initial weight pre-trained on the dataset.
The difference is that the teacher is frozen.
During the training stage, each batch contains pseudo frames $\{\boldsymbol{x}^p_i\}^{N_p}_{i=1}$ and real frames $\{\boldsymbol{x}^r_i\}^{N_r}_{i=1}$, where $N_p$ and $N_r$ are the numbers of pseudo frames and real frames.
The real frames $\{\boldsymbol{x}^r_i\}^{N_r}_{i=1}$ are used to supervise the student network as fully-supervised methods. 
For pseudo frames $\{\boldsymbol{x}^p_i\}^{N_p}_{i=1}$, they are first fed to the teacher network. The caption predictions $\{\boldsymbol{pred}^p_i\}_{i=1}^{N_p}$ of the teacher network will be used as pseudo caption labels of pseudo frames. 
To ensure quality, the pseudo caption labels with low confidence will be removed.
The pseudo caption labels and the ground truth labels $\{\boldsymbol{gt\_cap}^r_i\}_{i=1}^{N_r}$ are used to supervise the predicted captions for pseudo frames and real frames in the student network, respectively.

\subsection{VoxelCapNet}
\label{sec:voxel-caption-network}
\vspace{-1mm}
In this paper, we present the first voxel-based 3D dense caption network, VoxelCapNet.
We first rethink the design of the backbone and detection head, then adapt the caption head to the voxel-based network to build the VoxelCapNet.

\subsubsection{Backbone.}
We first apply voxelization on input point cloud $P$ to generate sparse voxels 
$V=\left[p_v; f_v\right] \in \mathbb{R}^{N \times \left(3 + D\right)}$, 
where $p_{v} \in \mathbb{R}^{N\times 3}$ is the discrete coordinates of each voxel, 
and $f_{v} \in \mathbb{R}^{N\times D}$ is the corresponding features.
Then, we leverage a modern voxel-based backbone to supply rich features for the downstream head:

\vspace{-5mm}
\begin{equation}
\begin{aligned}
    V_s = 
    VoxelBackbone(V), 
    V_s \in \mathbb{R}^{M \times \left(3 + C\right)},
\vspace{-4mm}
\end{aligned}
\end{equation}

\noindent where $V_s$ is the voxel feature map whose resolution is 1/2 of $V$, $M$ is the number of voxels in $V_s$, $C$ is the channels of the output features.

\subsubsection{Detection Head.}
A powerful detection head can provide accurate bounding boxes and robust representations for each object, which is crucial for the caption head to generate high-quality descriptions. Recently, voxel-based detectors \cite{fcaf3d, cagroup3d} show impressive results. 
In our VoxelCapNet, we leverage the detection head of \cite{cagroup3d} to extract high-quality object features and generate accurate bounding boxes for the caption head.
Specifically, after obtaining the voxel features from the backbone, we apply sparse convolutions to prepare object features, which can be used for the detection head and caption head: 

\vspace{-4mm}
\begin{equation}
\begin{aligned}
    O_q = 
    SparseConv(V_s).
\vspace{-1mm}
\end{aligned}
\end{equation}

\noindent Then, the voxel-based detection head is used to regress bounding boxes $B$ and predict confidence scores $S$, which can be formulated as follows:

\vspace{-2mm}
\begin{equation}
\begin{aligned}
    B, S  = 
    VoxelDetHead(O_q), 
\vspace{-1mm}
\end{aligned}
\end{equation}

\noindent where M is the number of predicted boxes. 
The predicted boxes whose confidence scores $S$ are lower than a predefined threshold will be filtered. Non-maximum suppression (NMS) is further conducted to drop duplicate predictions. Then, we get the final predicted boxes $\overline{B}$.
Given a set of generated high-quality proposals $\overline{B}$, we design the object mask $M \in \mathbb{R}^{M}$:

\vspace{-4mm}
\begin{equation}
\begin{aligned}
    M_i = \left\{
        \begin{array}{ll} 
        1\ , \quad & if \quad B_i \in \overline{B} \\
        0\ , \quad & otherwise
        \end{array}
    \right\}
    ,i = 0,1,2,...,M-1.
\end{aligned}
\end{equation}

\noindent The object mask $M$ is used to select the corresponding object features $V_q$ of high-quality predicted bounding boxes, as shown in Equation \ref{eq:select}. Finally, we feed the contexture features $V_s$, query features $V_q$ and object proposals $\overline{B}$ to the caption head for caption generation.

\vspace{-2mm}
\begin{equation}
\label{eq:select}
\begin{aligned}
    V_q = 
    O_q \cdot M.
\end{aligned}
\end{equation}

\subsubsection{Caption Head.}
After equipping the voxel-based backbone and detection head, we further adapt the caption head of Vote2Cap-DETR \cite{vote2cap} to the voxel-based network architecture. Vote2Cap-DETR directly leverages learnable queries of the detection head as caption queries. In our method, we take object features $V_q$ filtered by confidences and NMS of the detection head as query features.
The voxel features $V_s$ from the backbone are used as surrounding contextual features. 
Then, we feed $V_q$ and $V_s$ to the caption head for caption generation, as shown in the bottom left of Figure \ref{fig:overview}. 
We follow the query definition of Vote2Cap-DETR \cite{vote2cap}, where the query is the reference caption tokens that replace their Start Of Seqenece(‘SOS’) prefix by $V_q$. We leverage the $V_q$'s k-nearest local surrounding features $V_s$ as key and value.
Experiment results show that our design can significantly improve the caption performance.

\subsubsection{Training Objective.}
The final training objective is comprised of the caption loss and detection loss on real frames and pseudo frames:

\vspace{-1mm}
\begin{equation}
\begin{aligned}
\mathcal{L} = 
    \mathcal{L}_{\text{cap}} 
        \left(\{\boldsymbol{x}^p_i\}^{N_p}_{i=1},
        \{\boldsymbol{pred}^p_i\}_{i=1}^{N_p} \right) + \\
    \alpha \mathcal{L}_{\text{det}} 
        \left(\{\boldsymbol{x}^p_i\}^{N_p}_{i=1},
        \{\boldsymbol{gt\_det}^p_i\}_{i=1}^{N_p} \right) + \\
    \mathcal{L}_{\text{cap}} 
        \left(\{\boldsymbol{x}^r_i\}^{N_r}_{i=1},
        \{\boldsymbol{gt\_cap}^r_i\}_{i=1}^{N_r} \right) + \\
    \alpha \mathcal{L}_{\text{det}} 
        \left(\{\boldsymbol{x}^r_i\}^{N_r}_{i=1},
        \{\boldsymbol{gt\_det}^r_i\}_{i=1}^{N_r} \right),
\vspace{-1mm}
\end{aligned}
\end{equation}

\noindent where $\alpha$ is set to 1 to maintain a similar magnitude of different losses, $\{\boldsymbol{gt\_det}^p_i\}_{i=1}^{N_p}$ and $\{\boldsymbol{gt\_det}^r_i\}_{i=1}^{N_r}$ is the ground-truth bounding boxes of pseudo frames and real frames.
Note that $\{\boldsymbol{gt\_det}^p_i\}_{i=1}^{N_p}$ is accessible because the ground-truth bounding box of each instance in its original frame can be re-used in the pseudo frame.
For the detection loss ${L}_{\text{det}}$, we follow the definition of \cite{cagroup3d}.
For the caption loss ${L}_{\text{cap}}$, we follow Vote2Cap-DETR\cite{vote2cap} to train with standard cross-entropy loss (MLE training) and finetune with \textbf{S}elf-\textbf{C}ritical \textbf{S}equence \textbf{T}raining (SCST).

\section{Experiments}
\label{sec:experiments}
\subsection{Datasets, Metrics, and Implementation Details}

\begin{table*}[t]
\vspace{-2mm}
    \caption{
    \textbf{Evaluating PVCap on ScanRefer\cite{scanrefer}.}
    We adopt the established evaluation protocol from Scan2Cap\cite{scan2cap} and compare our results separately using both MLE and SCST against existing 3D dense captioning methods.
    We provide evaluation results for PVCap with SparseUNet\cite{minkunet} and Swin3D\cite{swin3d} as backbone respectively, since additional 2D features introduce large memory overhead with Swin3D.
    The methods marked * are trained with extra data.
    }
    \vspace{-1mm}
    \centering
    \resizebox{\linewidth}{!}{
    \begin{tabular}{cccccccccccccccccccccc}
    \toprule
    \multirow{3}{*}{Method}            & \multirow{3}{*}{Loss} &  & \multicolumn{9}{c}{w/o additional 2D input}                                                                                                       &  & \multicolumn{9}{c}{w/ additional 2D input}                                                                                                \\
                                       &                                      &  & \multicolumn{4}{c}{IoU = 0.25}                                    &  & \multicolumn{4}{c}{IoU = 0.50}                                    &  & \multicolumn{4}{c}{IoU = 0.25}                                    &  & \multicolumn{4}{c}{IoU = 0.50}                                    \\ \cline{4-7} \cline{9-12} \cline{14-17} \cline{19-22} 
                                       &                                      &  & C              & B-4            & M              & R              &  & C              & B-4            & M              & R              &  & C              & B-4            & M              & R              &  & C              & B-4            & M              & R              \\ \hline
    Scan2Cap\cite{scan2cap}            & \multirow{15}{*}{MLE}                &  & 53.73          & 34.25          & 26.14          & 54.95          &  & 35.20          & 22.36          & 21.44          & 43.57          &  & 56.82          & 34.18          & 26.29          & 55.27          &  & 39.08          & 23.32          & 21.97          & 44.78          \\
    MORE\cite{more}                    &                                      &  & 58.89          & 35.41          & 26.36          & 55.41          &  & 38.98          & 23.01          & 21.65          & 44.33          &  & 62.91          & 36.25          & 26.75          & 56.33          &  & 40.94          & 22.93          & 21.66          & 44.42          \\
    SpaCap3d\cite{spacap3d}            &                                      &  & -              & -              & -              & -              &  & 42.76          & 25.38          & 22.84          & 45.66          &  & -              & -              & -              & -              &  & 44.02          & 25.26          & 22.33          & 45.36          \\
    REMAN\cite{reman}                  &                                      &  & -              & -              & -              & -              &  & -              & -              & -              & -              &  & 62.01          & 36.37          & 26.76          & 56.25          &  & 45.00          & 26.31          & 22.67          & 46.96          \\
    Contextual\cite{contextual}        &                                      &  & -              & -              & -              & -              &  & 42.77          & 23.60          & 22.05          & 45.13          &  & -              & -              & -              & -              &  & 46.11          & 25.47          & 22.64          & 45.96          \\
    UniT3D$^{*}$\cite{unit3d}          &                                      &  & -              & -              & -              & -              &  & -              & -              & -              & -              &  & -              & -              & -              & -              &  & 46.69          & 27.22          & 21.91          & 45.98          \\
    3DJCG\cite{3djcg}                  &                                      &  & 60.86          & 39.67          & 27.45          & 59.02          &  & 47.68          & 31.53          & 24.28          & 51.80          &  & 64.70          & 40.17          & 27.66          & 59.23          &  & 49.48          & 31.03          & 24.22          & 50.80          \\
    D3Net\cite{d3net}                  &                                      &  & -              & -              & -              & -              &  & -              & -              & -              & -              &  & -              & -              & -              & -              &  & 46.07          & 30.29          & 24.35          & 51.67          \\
    3D-VLP$^{*}$\cite{3d-vlp}          &                                      &  & 64.09          & 39.84          & 27.65          & 58.78          &  & 50.02          & 31.87          & 24.53          & 51.17          &  & 70.73          & 41.03          & 28.14          & 59.72          &  & 54.94          & 32.31          & 24.83          & 51.51          \\
    3D-VisTA$^{*}$\cite{3d-vista}      &                                      &  & 71.00          & 36.50          & 28.40          & 57.60          &  & 66.90          & 34.00          & 27.10          & 54.30          &  & -              & -              & -              & -              &  & -              & -              & -              & -              \\
    Vote2Cap-DETR\cite{vote2cap}       &                                      &  & 71.45          & 39.34          & 28.25          & 59.33          &  & 61.81          & 34.46          & 26.22          & 54.40          &  & 72.79          & 39.17          & 28.06          & 59.23          &  & 59.32          & 32.42          & 25.28          & 52.53          \\ 
    Vote2Cap-DETR++\cite{vote2cap++}   &                                      &  & 76.36          & \textbf{41.37} & 28.70          & 60.00          &  & 67.58          & 37.05          & 26.89          & 55.64          &  & \textbf{77.03} & 40.99          & 28.53          & 59.59          &  & 64.32          & 34.73          & 26.04          & 53.67          \\ 
    \cline{1-1}
    \multicolumn{1}{l}{\textbf{\textit{Ours:}}} \\
    PVCap-SparseUNet                   &                                      &  & 72.21          & 40.95          & 28.83          & 62.88          &  & 66.16          & 38.73          & 27.82          & \textbf{60.04} &  & 74.24          & \textbf{41.99} & \textbf{29.11} & \textbf{63.35} &  & \textbf{68.67} & \textbf{39.12} & \textbf{27.90} & \textbf{59.20} \\ 
    PVCap-Swin3D                       &                                      &  & \textbf{77.33} & 40.78          & \textbf{29.39} & \textbf{62.73} &  & \textbf{71.90} & \textbf{39.72} & \textbf{28.30} & 59.92          &  & -              & -              & -              & -              &  & -              & -              & -              & -              \\ \hline
    
    $\chi$-Trans2Cap\cite{x-trans2cap} & \multirow{9}{*}{SCST}                &  & 58.81          & 34.17          & 25.81          & 54.10          &  & 41.52          & 23.83          & 21.90          & 44.97          &  & 61.83          & 35.65          & 26.61          & 54.70          &  & 43.87          & 25.05          & 22.46          & 45.28          \\
    Scan2Cap\cite{scan2cap}            &                                      &  & -              & -              & -              & -              &  & -              & -              & -              & -              &  & -              & -              & -              & -              &  & 48.38          & 26.09          & 22.15          & 44.74          \\
    D3Net\cite{d3net}                  &                                      &  & -              & -              & -              & -              &  & -              & -              & -              & -              &  & -              & -              & -              & -              &  & 62.64          & 35.68          & 25.72          & 53.90          \\
    Contextual\cite{contextual}        &                                      &  & -              & -              & -              & -              &  & 50.29          & 25.64          & 22.57          & 44.71          &  & -              & -              & -              & -              &  & 54.30          & 27.24          & 23.30          & 45.81          \\
    Vote2Cap-DETR\cite{vote2cap}       &                                      &  & 84.15          & 42.51          & 28.47          & 59.26          &  & 73.77          & 38.21          & 26.64          & 54.71          &  & 86.28          & 42.64          & 28.27          & 59.07          &  & 70.63          & 35.69          & 25.51          & 52.28          \\ 
    Vote2Cap-DETR++\cite{vote2cap++}   &                                      &  & 88.28          & 44.07          & 28.75          & 59.89          &  & 78.16          & 39.72          & 26.94          & 55.52          &  & 88.56          & 43.30          & 28.64          & 59.19          &  & 74.44          & 37.18          & 26.20          & 53.30          \\ 
    \cline{1-1}
    \multicolumn{1}{l}{\textbf{\textit{Ours:}}} \\
    PVCap-SparseUNet                   &                                      &  & 88.87          & 45.94          & 28.98          & 63.77          &  & 79.64	         & 42.82	      & 27.65	       & \textbf{60.22} &  & \textbf{88.93} & \textbf{45.56} & \textbf{29.23} & \textbf{63.72} &  & \textbf{82.52} & \textbf{42.73} & \textbf{28.00} & \textbf{60.11} \\ 
    PVCap-Swin3D                       &                                      &  & \textbf{96.11} & \textbf{45.81} & \textbf{29.47} & \textbf{62.98} &  & \textbf{89.57} & \textbf{43.16} & \textbf{28.35} & 59.97          &  & -              & -              & -              & -              &  & -              & -              & -              & -              \\ \bottomrule
    \end{tabular}
    }
    \vspace{-4mm}
    \label{exp:comparison on scanrefer}
\end{table*}

\begin{table}[!t]
    \caption{
    \textbf{Evaluating PVCap on Nr3D\cite{referit3d}.}
    Likewise, our method is trained and evaluated on the Nr3D dataset, 
    outperforming previous approaches in both MLE training and SCST finetuning.
    }\label{exp:comparison on nr3d}
    \centering
    \vspace{-1mm}
    \resizebox{.98\linewidth}{!}{
        \begin{tabular}{cccccc}
            \toprule
            Method                            & Loss   & C@0.5           & B-4@0.5           & M@0.5           & R@0.5           \\ \hline
            Scan2Cap\cite{scan2cap}           & \multirow{10}{*}{MLE} & 27.47           & 17.24             & 21.80           & 49.06           \\
            SpaCap3d\cite{spacap3d}           &                       & 33.71           & 19.92             & 22.61           & 50.50           \\
            D3Net\cite{d3net}                 &                       & 33.85           & 20.70             & 23.13           & 53.38           \\
            REMAN\cite{reman}                 &                       & 34.81           & 20.37             & 23.01           & 50.99           \\
            Contextual\cite{contextual}       &                       & 35.26           & 20.42             & 22.77           & 50.78           \\
            3DJCG\cite{3djcg}                 &                       & 38.06           & 22.82             & 23.77           & 52.99           \\
            Vote2Cap-DETR\cite{vote2cap}      &                       & 43.84           & 26.68             & 25.41           & 54.43           \\ 
            Vote2Cap-DETR++\cite{vote2cap++}  &                       & 47.08           & 27.70             & 25.44           & 55.22           \\
            \cline{1-1}
            \multicolumn{1}{l}{\textbf{\textit{Ours:}}} \\
            PVCap-Swin3D                      &                       & \textbf{53.31}  & \textbf{30.14}    & \textbf{27.61}  & \textbf{59.98}  \\
            \hline
            $\chi$-Tran2Cap\cite{x-trans2cap} & \multirow{8}{*}{SCST} & 33.62           & 19.29             & 22.27           & 50.00           \\
            Contextual\cite{contextual}       &                       & 37.37           & 20.96             & 22.89           & 51.11           \\
            D3Net\cite{d3net}                 &                       & 38.42           & 22.22             & 24.74           & 54.37           \\
            Vote2Cap-DETR\cite{vote2cap}      &                       & 45.53           & 26.88             & 25.43           & 54.76           \\ 
            Vote2Cap-DETR++\cite{vote2cap++}  &                       & 47.62           & 28.41             & 25.63           & 54.77           \\
            \cline{1-1}
            \multicolumn{1}{l}{\textbf{\textit{Ours:}}} \\
            PVCap-Swin3D                      &                       & \textbf{61.61}  & \textbf{33.46}    & \textbf{27.49}  & \textbf{60.55}  \\
            \bottomrule
            \end{tabular}
    }
\vspace{-6mm}
\end{table}

\subsubsection{Datasets.}
We performed experiments on two commonly used datasets: \emph{i.e.}, ScanRefer \cite{scanrefer} and Nr3D \cite{referit3d}. 
Both datasets are derived from 3D scenes within ScanNet \cite{scannet}, which contains 1201 indoor 3D scenes for training and 312 for validation.
ScanRefer consists of 36665 human-annotated natural language descriptions for training, covering 7875 objects from 562 3D scenes. For evaluation, it provides 9508 sentences describing 2068 objects from 141 3D scenes.
Nr3D contains 32919 language annotations describing 4664 objects from 511 3D scenes for training. For evaluation, it offers 8584 annotations for 1214 objects from 130 3D scenes.

\subsubsection{Evaluation Metrics.}
According to \cite{scan2cap}, we combine standard image captioning metrics with Intersection-over-Union (IoU) 
scores between predicted bounding boxes and the target bounding boxes for evaluation, in order to jointly 
measure the quality of the generated descriptions and the accuracy of the detected bounding boxes.
We adopt the m@kIoU metric:
\vspace{-2mm}
\begin{equation}
  m@k\text{IoU}=\frac{1}{N}\sum^{N}_{i=0}m_{i}u_{i},
  \vspace{-1mm}
\end{equation}
where $N$ is the number of ground truth instances, and $u_i \in \{0,1\}$ is set to $1$ if the IoU 
score for the $i^{th}$ box is greater than $k$, otherwise 0. $m$ represents the captioning metrics 
CIDEr\cite{cider}, BLEU-4\cite{bleu}, METEOR\cite{meteor} and ROUGE\cite{rouge} (C, B-4, M, and R).
We follow a standard evaluation protocol~\cite{qi2019deep} using mean Average Precision (mAP) 
under IoU thresholds of 0.25 and 0.50 for 3D object detection. 

\vspace{-2mm}
\subsubsection{Implementation Details.}
\vspace{-1mm}
Our approach is based on the Pointcept codebase. 
For all experiments set below, we use an AdamW optimizer with 1e-4 weight decay and a OneCycle scheduler with a maximum learning rate of 5e-4.
We conduct training with a batch size of 8 and evaluate the model after every 10 epochs with a batch size of 4, performing on 4 Nvidia A100 GPUs.
Under the MLE training, we first pre-train a VoxelCapNet for 300 epochs.
Then, we load the pre-trained weight to both the teacher and student in Figure \ref{fig:overview} 
and train the student with PseudoCap for another 300 epochs.
Here, we freeze the weights of the teacher and utilize caption predictions 
of the teacher as caption pseudo labels to train the student.
When training the model with PseudoCap, we randomly select samples in a batch with 
a probability of p (p=0.5 by default) for PseudoCap data augmentation.
Given a well-trained model, we further perform Self-Critical Sequence Training (SCST) 
following \cite{scst, vote2cap}. Specifically, we freeze all the weight of the model 
except for the caption head and fine-tune 
for 100 epochs.

\vspace{-1mm}
\subsection{Quantitative Comparisons}
We evaluate our PVCap against previous approaches on two commonly used datasets, ScanRefer and Nr3D.
C, B-4, M, and R are respectively utilized under IoU thresholds of 0.25, 0.5 for ScanRefer 
in Table \ref{exp:comparison on scanrefer} and 0.5 for Nr3D in Table \ref{exp:comparison on nr3d}.
For ScanRefer in Table \ref{exp:comparison on scanrefer}, we provide the experimental results trained  
with point clouds that with/without additional 2D features, respectively.
We examine our PVCap with two different backbones: SparseUNet\cite{minkunet} and Swin3D\cite{swin3d}.
Because training PVCap-Swin3D with additional 2D features leads to large memory consumption, 
we only provide the results of PVCap-SparseUNet with additional 2D features.  
Due to the significant effect of the SCST tuning process on captioning performance, 
a separate comparison with previous methods is also presented in Table \ref{exp:comparison on scanrefer} following \cite{vote2cap}.

As shown in Table \ref{exp:comparison on scanrefer}, our PVCap surpasses all state-of-the-art methods. 
Under the MLE training, PVCap-Swin3D achieves 71.90\% C@0.5IoU without additional 2D inputs, 
4.32\% higher than the current best result reported by Vote2Cap-DETR++\cite{vote2cap++}.
Besides, PVCap-SparseUNet reaches 68.67\% (+4.35\%) C@0.5IoU with additional 2D inputs. 
In addition, Table \ref{exp:map analysis} shows that our PVCap-Swin3D outperforms 
Vote2Cap-DETR++\cite{vote2cap++} by 6.33\% mAP@0.5 on the detection task.
Following the SCST tuning, PVCap-Swin3D trained without additional 2D inputs and PVCap-SparseUNet 
trained with additional 2D inputs surpasses the previous best method on C@0.5IoU by 11.41\% and 8.08\% respectively. 
The performance comparison with other approaches on Nr3D is reported in Table \ref{exp:comparison on nr3d}.
Under MLE training, our PVCap-Swin3D reaches 53.31\% C@0.5IoU, which surpasses the current best result by 6.23\%.
Under SCST, PVCap-Swin3D performs 61.61\% C@0.5IoU, achieving a significant improvement of 13.99\%.
The significant improvement to previous methods on ScanRefer and Nr3D strongly demonstrates the effectiveness of our method.

\begin{table}[!t]  
    \caption{
    \textbf{Performance of 3D object detection.}
    We compare the 3D object detection performance of our proposed PVCap with Vote2Cap-DETR\cite{vote2cap} and Vote2Cap-DETR++\cite{vote2cap++} on ScanRefer\cite{scanrefer} without additional 2D features, 
    using SparseUNet\cite{minkunet} and Swin3D\cite{swin3d} as backbone, respectively.
    }\label{exp:map analysis}
    \centering
    \vspace{-1mm}
    \resizebox{1.02\linewidth}{!}{
    \renewcommand\arraystretch{1.16}
    \begin{tabular}{ccccccc}
    \toprule
    ID  & Method                           &  & mAP@0.25       & AR@0.25        & mAP@0.5        & AR@0.5         \\ \hline
    (1) & Vote2Cap\cite{vote2cap}     &  & 69.61          & 87.20          & 52.13          & 69.12          \\ 
    (2) & Vote2Cap++\cite{vote2cap++} &  & 70.52          & 85.64          & 55.48          & 70.89          \\ \cline{1-2}
    (3) & PVCap-SparseUNet                 &  & 71.75          & 90.38          & 59.87          & 76.41          \\
    (4) & PVCap-Swin3D                     &  & \textbf{75.65} & \textbf{90.48} & \textbf{61.81} & \textbf{77.00} \\ \bottomrule 
    \end{tabular}
    }
\vspace{-3mm}
\end{table}

\begin{table}[!t]
    \caption{
    \textbf{Ablation study on each component of our PVCap.}
    Based on our baseline Vote2cap-DETR\cite{vote2cap}, we add VoxelCapNet (\cref{sec:voxel-caption-network}) and PseudoCap (\cref{sec:pseudo-caption}) gradually to validate the effectiveness of them.
    }
    \centering
    \vspace{-1mm}
    \resizebox{1.02\linewidth}{!}{
    \renewcommand\arraystretch{1.16}
    \begin{tabular}{ccccccccc}
    \toprule
    ID  & Baseline   & VoxelCapNet & PseudoCap  & & C@0.5          & B-4@0.5        & M@0.5          & R@0.5          \\ \hline
    (1) & \checkmark &             &            & & 61.81          & 34.46          & 26.22          & 54.40          \\
    (2) & \checkmark & \checkmark  &            & & 69.59          & 39.24          & 28.05          & \textbf{60.04} \\
    (3) & \checkmark & \checkmark  & \checkmark & & \textbf{71.90} & \textbf{39.72} & \textbf{28.30} & 59.92          \\ \bottomrule 
    \end{tabular}
    }
    \label{exp:components}
\vspace{-3mm}
\end{table}

\begin{table}[!t]   
    \caption{
    \textbf{Ablation study on different backbones.}
    We conduct 300-epoch training for VoxelCapNet with different backbones and present the evaluation results below. We also report the average execution time for each backbone.
    }\label{exp:backbone analysis}
    \centering
    \vspace{-1mm}
    \resizebox{1.02\linewidth}{!}{
    \renewcommand\arraystretch{1.16}
    \begin{tabular}{cccccccc}
    \toprule
    ID  & Backbone                    &  & C@0.5          & B-4@0.5        & M@0.5          & R@0.5          & Time        \\ \hline
    (1) & SparseUNet\cite{minkunet} &  & 66.35          & 39.35 & 28.01          & 60.32 & 41 ms \\ 
    (2) & PTV2\cite{ptv2}      &  & 61.94          & 36.42          & 26.78          & 57.32          & 118 ms         \\ 
    (3) & Swin3D\cite{swin3d} &  & 69.59 & 39.24          & 28.05 & 60.04          & 390 ms        \\ \bottomrule 
    \end{tabular}
    }
   \vspace{-4mm}
\end{table}

\subsection{Ablation Studies}
We perform extensive ablation studies to understand the effectiveness of each component of our PVCap.
Without further specification, all experiments are conducted on ScanRefer by PVCap-Swin3D without additional 2D features.

\subsubsection{Effect of VoxelCapNet.}
We first investigate the effect of VoxelCapNet.
The experiment (1) of Table \ref{exp:components} is our baseline Vote2Cap-DETR\cite{vote2cap}.
It adopts the 3DETR architecture\cite{3detr} with vote query generation and 
uses output features from its encoder and decoder as surrounding contextual features ($V_s$) and object features ($V_q$), respectively. 
In our method, we replace the encoder-decoder architecture with the VoxelCapNet.
The VoxelCapNet can extract more informative contextual features $V_s$ from its powerful backbone 
and more accurate object features $V_q$ from its strong detection head.
Comparing experiments (1) and (2) of Table \ref{exp:components}, we can find that our VoxelCapNet achieves an 
impressive performance of 69.59\% C@0.5IoU, exceeding the baseline by 7.78\%. 
This encouraging result emphasizes the effectiveness of our VoxelCapNet and 
shows that strong network architecture is crucial for high-performance 3D dense captioning.
Besides, the strong result achieved by VoxelCapNet can also make it a good baseline for further investigation on 3D dense captioning.

\begin{table}[!t]   
    \caption{
    \textbf{Ablation study on the probability of PseudoCap.}
    The best result is achieved with a probability of 0.50.
    }\label{exp:aug analysis}
    \centering
    \vspace{-1mm}
    \resizebox{.85\linewidth}{!}{
    \begin{tabular}{ccccccc}
    \toprule
    ID  & Prob. &  & C@0.5          & B-4@0.5        & M@0.5          & R@0.5          \\ \hline
    (1) & 0     &  & 66.92          & 38.34          & 27.72          & 59.14          \\ 
    (2) & 0.25  &  & 69.20          & 37.33          & 28.14          & 59.15          \\ 
    (3) & 0.50  &  & \textbf{71.90} & \textbf{39.72} & \textbf{28.30} & \textbf{59.92} \\ 
    (4) & 0.75  &  & 69.79          & 38.55          & 28.21          & 59.79          \\ \bottomrule
    \end{tabular}
    }
   \vspace{-2mm}
\end{table}

\begin{table}[!t]   
    \caption{
        \textbf{Results of training for different epochs.}
        It can be seen that training the model for 300 epochs yields the best performance.
        }\label{exp:epoch analysis}
    \centering
    \vspace{-1mm}
    \resizebox{.98\linewidth}{!}{
    \begin{tabular}{cccccccc}
        \toprule
        ID  & Training Epochs &  & C@0.5          & B-4@0.5        & M@0.5          & R@0.5          \\ \hline
        (1) & 150             &  & 65.55          & 39.12          & 27.69          & 59.89          \\ 
        (2) & 300             &  & \textbf{69.59} & \textbf{39.24} & \textbf{28.05} & \textbf{60.04} \\
        (3) & 600             &  & 67.94          & 38.93          & 28.00          & 59.54          \\ 
        (4) & 900             &  & 67.10          & 37.83          & 27.75          & 59.03          \\ \bottomrule
        \end{tabular}
    }
   \vspace{-4mm}
\end{table}

\vspace{-1mm}
\subsubsection{Effect of PseudoCap.}
We conduct an ablation study to explore the effect of PseudoCap, as shown in the experiment (2) of Table \ref{exp:components}.
PseudoCap mixes instances cropped from different frames and generates 
diverse pseudo frames to provide a strong data augmentation for 3D dense captioning.
Specifically, we load the weight of experiment (2) to initialize the teacher and student of the experiment (3), 
and then we train the model with PseudoCap with an extra 300 epochs.
By comparing experiments (2) and (3), we find that our PseudoCap can also bring considerable improvement.
In particular, the performance is boosted to 71.90\% C@0.5IoU, which is 2.31\% higher than experiment (2).
Moreover, we provide an ablation to investigate if the improvement brought by PseudoCap results from the extra training epochs.
In experiment (1) of Table \ref{exp:aug analysis}, we perform PseudoCap with 
a probability of 0, which means that we do not use PseudoCap for extra training.
Comparing experiment (1) of Table \ref{exp:aug analysis} with experiment (2) of Table \ref{exp:components},
we find that training more epochs even leads to performance degradation, 
which can be caused by overfitting the dataset after training too many epochs.
This also confirms the effectiveness of our PseudoCap.

\begin{figure*}[t]
	\begin{center}
		\setlength{\fboxrule}{0pt}
		\fbox{\includegraphics[width=1.00\textwidth]{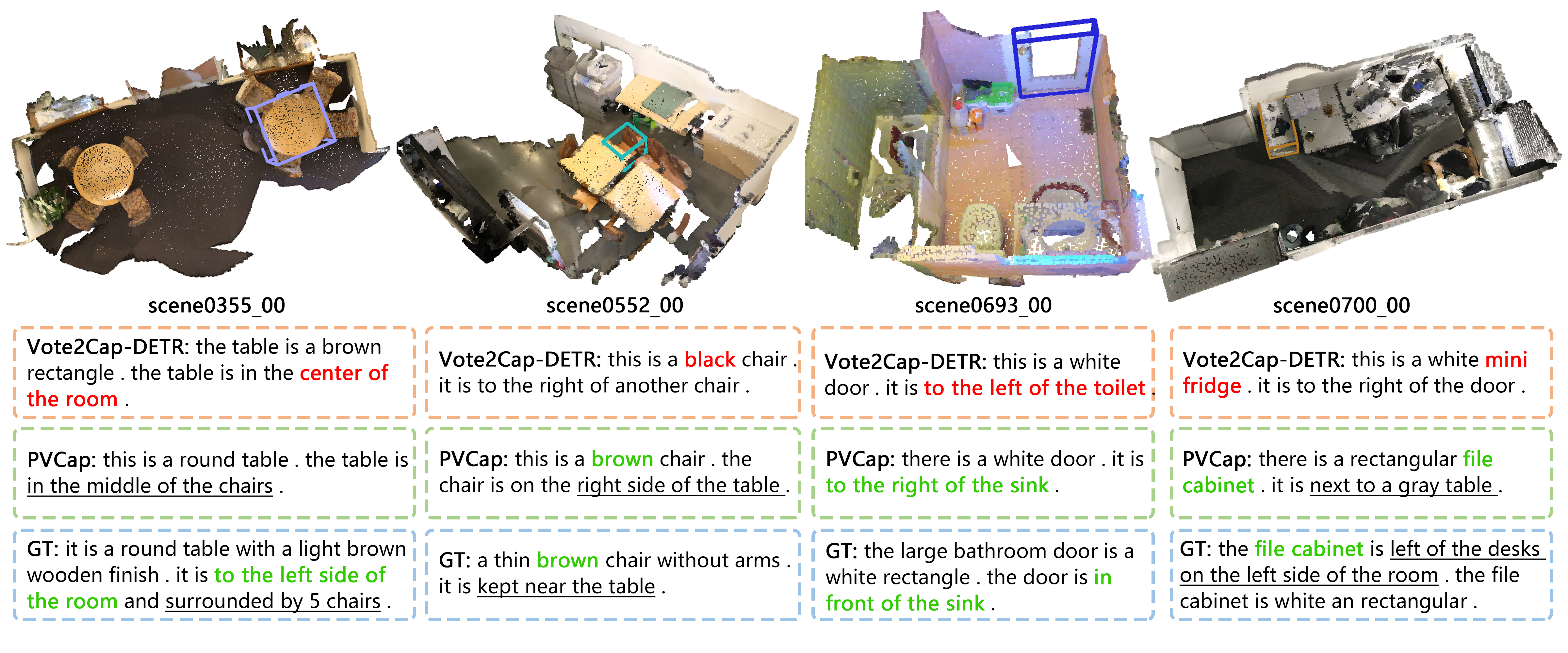}}
	\end{center}
	\vspace{-9mm}
	\caption{\textbf{Qualitative Comparisons.}
 We compare visualization results with Vote2Cap-DETR\cite{vote2cap} 
 and showcase the ground-truth captions. 
 In each scene, the wrong descriptions are marked in red and the corresponding correct descriptions are in green. Besides, we underline the surrounding information to exhibit the ability of our PVCap on environment description.
 }
	\label{fig:visual}
	\vspace{-4mm}
\end{figure*}

\subsubsection{Ablation on Backbone for VoxelCapNet.}
In this paragraph, we ablate the effect of different backbones on the performance of 3D dense captioning.
Considering the impressive results achieved by SparseUnet\cite{minkunet}, PTV2\cite{ptv2} and Swin3D\cite{swin3d} 
on 3D semantic segmentation, we utilize them as the backbone of VoxelCapNet.
Here, SparseUNet is the re-implementation of MinkowskiNet \cite{minkunet} by the Pointcept codebase.
As shown in Table \ref{exp:backbone analysis}, Swin3D performs the best (69.59\% C@0.5IoU) among the three backbones 
while it consumes more computation cost (391 ms). SparseUNet achieves a moderate performance (66.35\% C@0.5IoU) and 
a low inference latency (41 ms), which is more practical in real-world applications.

\subsubsection{Ablation on the Probability of PseudoCap.}
When training PseudoCap, we randomly select some real frames from a batch with a probability 
and transfer them to pseudo frames. Here, we provide the ablation study on the probability of PseudoCap.
As summarized in Table \ref{exp:aug analysis}, we utilize 4 different probabilities for PseudoCap.
A probability of 0 means that we do not perform PseudoCap. Experiment (1) in Table \ref{exp:aug analysis} 
indicates that training the model without PseudoCap only achieves 66.92\% in C@0.5IoU.
When we use a probability of 0.25 and 0.5, the performance increases to 69.20\% and 71.90\%, as shown in experiments (2) and (3).
However, we notice a decline of 2.11\% C@0.5IoU when the probability comes to 0.75.
The reason can be that a large probability for PseudoCap results in a small number of real frames, 
which can provide more accurate supervision than pseudo frames and is also essential.

\subsubsection{Abaltion on Training Epochs.}
The evaluation results of our VoxelCapNet trained with different epochs are exhibited in Table \ref{exp:epoch analysis}.
Our method shows a faster convergence than Vote2Cap-DETR\cite{vote2cap}, which needs 720 epochs.
Even trained with 150 epochs, our method can achieve a higher performance, which 
surpasses the Vote2Cap-DETR by 3.74\% C@0.5IoU.
After training for 300 epochs, our method achieves the best performance, 69.59\% C@0.5IoU, which 
outperforms the Vote2Cap-DETR by 7.78\% C@0.5IoU.
Experiments (3) and (4) suggest that the performance will drop when training with 600 and 900 epochs, 
which can be because more training epochs may cause overfitting.

\vspace{0.5mm}
\subsection{Qualitative Comparison}
We compare qualitative results with Vote2Cap-DETR\cite{vote2cap} and the ground truth in Figure 
\ref{fig:visual}. 
The wrong descriptions are marked in red and the corresponding correct ones are in green.
Apart from obtaining correct categories for objects, our PVCap can provide more accurate descriptions of the color.
Moreover, our method can also generate more comprehensive and more accurate captions about the surrounding objects, 
exhibiting superior ability in environment description.
The impressive visualizations confirm the effectiveness of our method again.



\section{Conclusion}
\label{sec:conclusion}
This work proposes a top-performing 3D dense captioning framework dubbed PVCap. We propose a novel data augmentation, PseudoCap, which mixes instances from different scenes to produce pseudo frames with diverse spatial layouts. We also present a strong baseline, termed VoxelCapNet, for 3D dense captioning. VoxelCapNet is based on voxel features and exhibits impressive performance on 3D dense captioning tasks. Our method attains a new state-of-the-art on ScanRefer and Nr3D benchmarks and extensive experiments validate the effectiveness of our approach.

{\small
\bibliographystyle{ieee_fullname}
\bibliography{egbib}

@String(ECCV= {Eur. Conf. Comput. Vis.})

@String(IJCAI = {IJCAI})

@String(ECCV  = {ECCV})

@inproceedings{vote2cap,
  title={End-to-end 3d dense captioning with vote2cap-detr},
  author={Chen, Sijin and Zhu, Hongyuan and Chen, Xin and Lei, Yinjie and Yu, Gang and Chen, Tao},
  booktitle={Proceedings of the IEEE/CVF Conference on Computer Vision and Pattern Recognition},
  pages={11124--11133},
  year={2023}
}

@misc{vote2cap++,
      title={Vote2Cap-DETR++: Decoupling Localization and Describing for End-to-End 3D Dense Captioning}, 
      author={Sijin Chen and Hongyuan Zhu and Mingsheng Li and Xin Chen and Peng Guo and Yinjie Lei and Gang Yu and Taihao Li and Tao Chen},
      year={2023},
      eprint={2309.02999},
      archivePrefix={arXiv},
      primaryClass={cs.CV}
}

@inproceedings{sad,
  title={Learning lightweight lane detection \uppercase{cnn}s by self attention distillation},
  author={Hou, Yuenan and Ma, Zheng and Liu, Chunxiao and Loy, Chen Change},
  booktitle={IEEE International Conference on Computer Vision},
  pages={1013--1021},
  year={2019}
}

@article{moe3d,
  title={MoE3D: Mixture of Experts meets Multi-Modal 3D Understanding},
  author={Li, Yu and Hou, Yuenan and Wei, Yingmei and Zhu, Xinge and Ma, Yuexin and Shao, Wenqi and Guo, Yanming},
  journal={arXiv preprint arXiv:2511.22103},
  year={2025}
}

@article{sega3d,
  title={Segment and Select: Vision-Language Segmentation in 3D Scenarios},
  author={Chen, Yulin and Zhong, Zhihang and Hou, Yuenan},
  journal={arXiv preprint arXiv:2606.10594},
  year={2026}
}

@inproceedings{taseg,
  title={Taseg: Temporal aggregation network for lidar semantic segmentation},
  author={Wu, Xiaopei and Hou, Yuenan and Huang, Xiaoshui and Lin, Binbin and He, Tong and Zhu, Xinge and Ma, Yuexin and Wu, Boxi and Liu, Haifeng and Cai, Deng and others},
  booktitle={IEEE Conference on Computer Vision and Pattern Recognition},
  pages={15311--15320},
  year={2024}
}

@article{nerf-det++,
  title={Nerf-det++: Incorporating semantic cues and perspective-aware depth supervision for indoor multi-view 3d detection},
  author={Huang, Chenxi and Hou, Yuenan and Ye, Weicai and Huang, Di and Huang, Xiaoshui and Lin, Binbin and Cai, Deng},
  journal={IEEE Transactions on Image Processing},
  year={2025},
  publisher={IEEE}
}

@inproceedings{pvkd,
    author    = {Hou, Yuenan and Zhu, Xinge and Ma, Yuexin and Loy, Chen Change and Li, Yikang},
    title     = {Point-to-{V}oxel {K}nowledge {D}istillation for {L}i{DAR} {S}emantic {S}egmentation},
    booktitle = {IEEE {C}onference on {C}omputer {V}ision and {P}attern {R}ecognition},
    year      = {2022},
    pages     = {8479-8488}
}

@inproceedings{scan2cap,
  title={Scan2cap: Context-aware dense captioning in rgb-d scans},
  author={Chen, Zhenyu and Gholami, Ali and Nie{\ss}ner, Matthias and Chang, Angel X},
  booktitle={Proceedings of the IEEE/CVF conference on computer vision and pattern recognition},
  pages={3193--3203},
  year={2021}
}

@inproceedings{referit3d,
  title={Referit3d: Neural listeners for fine-grained 3d object identification in real-world scenes},
  author={Achlioptas, Panos and Abdelreheem, Ahmed and Xia, Fei and Elhoseiny, Mohamed and Guibas, Leonidas},
  booktitle={Computer Vision--ECCV 2020: 16th European Conference, Glasgow, UK, August 23--28, 2020, Proceedings, Part I 16},
  pages={422--440},
  year={2020},
  organization={Springer}
}

@inproceedings{scanrefer,
  title={Scanrefer: 3d object localization in rgb-d scans using natural language},
  author={Chen, Dave Zhenyu and Chang, Angel X and Nie{\ss}ner, Matthias},
  booktitle={European conference on computer vision},
  pages={202--221},
  year={2020},
  organization={Springer}
}

@inproceedings{more,
  title={More: Multi-order relation mining for dense captioning in 3d scenes},
  author={Jiao, Yang and Chen, Shaoxiang and Jie, Zequn and Chen, Jingjing and Ma, Lin and Jiang, Yu-Gang},
  booktitle={European Conference on Computer Vision},
  pages={528--545},
  year={2022},
  organization={Springer}
}

@article{reman,
  title={Complete 3D Relationships Extraction Modality Alignment Network for 3D Dense Captioning},
  author={Mao, Aihua and Yang, Zhi and Chen, Wanxin and Yi, Ran and Liu, Yong-jin},
  journal={IEEE Transactions on Visualization and Computer Graphics},
  year={2023},
  publisher={IEEE}
}

@article{spacap3d,
  title={Spatiality-guided transformer for 3d dense captioning on point clouds},
  author={Wang, Heng and Zhang, Chaoyi and Yu, Jianhui and Cai, Weidong},
  journal={arXiv preprint arXiv:2204.10688},
  year={2022}
}

@inproceedings{d3net,
  title={D 3 Net: A Unified Speaker-Listener Architecture for 3D Dense Captioning and Visual Grounding},
  author={Chen, Dave Zhenyu and Wu, Qirui and Nie{\ss}ner, Matthias and Chang, Angel X},
  booktitle={European Conference on Computer Vision},
  pages={487--505},
  year={2022},
  organization={Springer}
}

@inproceedings{3d-vlp,
  title={Context-aware Alignment and Mutual Masking for 3D-Language Pre-training},
  author={Jin, Zhao and Hayat, Munawar and Yang, Yuwei and Guo, Yulan and Lei, Yinjie},
  booktitle={Proceedings of the IEEE/CVF Conference on Computer Vision and Pattern Recognition},
  pages={10984--10994},
  year={2023}
}

@inproceedings{unit3d,
  title={Unit3d: A unified transformer for 3d dense captioning and visual grounding},
  author={Chen, Zhenyu and Hu, Ronghang and Chen, Xinlei and Nie{\ss}ner, Matthias and Chang, Angel X},
  booktitle={Proceedings of the IEEE/CVF International Conference on Computer Vision},
  pages={18109--18119},
  year={2023}
}

@inproceedings{3djcg,
  title={3djcg: A unified framework for joint dense captioning and visual grounding on 3d point clouds},
  author={Cai, Daigang and Zhao, Lichen and Zhang, Jing and Sheng, Lu and Xu, Dong},
  booktitle={Proceedings of the IEEE/CVF Conference on Computer Vision and Pattern Recognition},
  pages={16464--16473},
  year={2022}
}

@inproceedings{3d-vista,
  title={3d-vista: Pre-trained transformer for 3d vision and text alignment},
  author={Zhu, Ziyu and Ma, Xiaojian and Chen, Yixin and Deng, Zhidong and Huang, Siyuan and Li, Qing},
  booktitle={Proceedings of the IEEE/CVF International Conference on Computer Vision},
  pages={2911--2921},
  year={2023}
}

@misc{scannet,
      title={ScanNet: Richly-annotated 3D Reconstructions of Indoor Scenes}, 
      author={Angela Dai and Angel X. Chang and Manolis Savva and Maciej Halber and Thomas Funkhouser and Matthias Nießner},
      year={2017},
      eprint={1702.04405},
      archivePrefix={arXiv},
      primaryClass={cs.CV}
}

@misc{cider,
      title={CIDEr: Consensus-based Image Description Evaluation}, 
      author={Ramakrishna Vedantam and C. Lawrence Zitnick and Devi Parikh},
      year={2015},
      eprint={1411.5726},
      archivePrefix={arXiv},
      primaryClass={cs.CV}
}

@inproceedings{meteor,
  title={METEOR: An automatic metric for MT evaluation with improved correlation with human judgments},
  author={Banerjee, Satanjeev and Lavie, Alon},
  booktitle={Proceedings of the acl workshop on intrinsic and extrinsic evaluation measures for machine translation and/or summarization},
  pages={65--72},
  year={2005}
}

@inproceedings{bleu,
  title={Bleu: a method for automatic evaluation of machine translation},
  author={Papineni, Kishore and Roukos, Salim and Ward, Todd and Zhu, Wei-Jing},
  booktitle={Proceedings of the 40th annual meeting of the Association for Computational Linguistics},
  pages={311--318},
  year={2002}
}

@inproceedings{rouge,
  title={Rouge: A package for automatic evaluation of summaries},
  author={Lin, Chin-Yew},
  booktitle={Text summarization branches out},
  pages={74--81},
  year={2004}
}

@inproceedings{x-trans2cap,
  title={X-trans2cap: Cross-modal knowledge transfer using transformer for 3d dense captioning},
  author={Yuan, Zhihao and Yan, Xu and Liao, Yinghong and Guo, Yao and Li, Guanbin and Cui, Shuguang and Li, Zhen},
  booktitle={Proceedings of the IEEE/CVF Conference on Computer Vision and Pattern Recognition},
  pages={8563--8573},
  year={2022}
}

@inproceedings{dou2022empirical,
  title={An empirical study of training end-to-end vision-and-language transformers},
  author={Dou, Zi-Yi and Xu, Yichong and Gan, Zhe and Wang, Jianfeng and Wang, Shuohang and Wang, Lijuan and Zhu, Chenguang and Zhang, Pengchuan and Yuan, Lu and Peng, Nanyun and others},
  booktitle={Proceedings of the IEEE/CVF Conference on Computer Vision and Pattern Recognition},
  pages={18166--18176},
  year={2022}
}

@inproceedings{jiang2018recurrent,
  title={Recurrent fusion network for image captioning},
  author={Jiang, Wenhao and Ma, Lin and Jiang, Yu-Gang and Liu, Wei and Zhang, Tong},
  booktitle={Proceedings of the European conference on computer vision (ECCV)},
  pages={499--515},
  year={2018}
}

@inproceedings{li2022blip,
  title={Blip: Bootstrapping language-image pre-training for unified vision-language understanding and generation},
  author={Li, Junnan and Li, Dongxu and Xiong, Caiming and Hoi, Steven},
  booktitle={International Conference on Machine Learning},
  pages={12888--12900},
  year={2022},
  organization={PMLR}
}

@inproceedings{yang2022vision,
  title={Vision-language pre-training with triple contrastive learning},
  author={Yang, Jinyu and Duan, Jiali and Tran, Son and Xu, Yi and Chanda, Sampath and Chen, Liqun and Zeng, Belinda and Chilimbi, Trishul and Huang, Junzhou},
  booktitle={Proceedings of the IEEE/CVF Conference on Computer Vision and Pattern Recognition},
  pages={15671--15680},
  year={2022}
}

@inproceedings{chen2020uniter,
  title={Uniter: Universal image-text representation learning},
  author={Chen, Yen-Chun and Li, Linjie and Yu, Licheng and El Kholy, Ahmed and Ahmed, Faisal and Gan, Zhe and Cheng, Yu and Liu, Jingjing},
  booktitle={European conference on computer vision},
  pages={104--120},
  year={2020},
  organization={Springer}
}

@inproceedings{hu2022scaling,
  title={Scaling up vision-language pre-training for image captioning},
  author={Hu, Xiaowei and Gan, Zhe and Wang, Jianfeng and Yang, Zhengyuan and Liu, Zicheng and Lu, Yumao and Wang, Lijuan},
  booktitle={Proceedings of the IEEE/CVF conference on computer vision and pattern recognition},
  pages={17980--17989},
  year={2022}
}

@article{li2021align,
  title={Align before fuse: Vision and language representation learning with momentum distillation},
  author={Li, Junnan and Selvaraju, Ramprasaath and Gotmare, Akhilesh and Joty, Shafiq and Xiong, Caiming and Hoi, Steven Chu Hong},
  journal={Advances in neural information processing systems},
  volume={34},
  pages={9694--9705},
  year={2021}
}

@inproceedings{li2020oscar,
  title={Oscar: Object-semantics aligned pre-training for vision-language tasks},
  author={Li, Xiujun and Yin, Xi and Li, Chunyuan and Zhang, Pengchuan and Hu, Xiaowei and Zhang, Lei and Wang, Lijuan and Hu, Houdong and Dong, Li and Wei, Furu and others},
  booktitle={Computer Vision--ECCV 2020: 16th European Conference, Glasgow, UK, August 23--28, 2020, Proceedings, Part XXX 16},
  pages={121--137},
  year={2020},
  organization={Springer}
}

@article{su2019vl,
  title={Vl-bert: Pre-training of generic visual-linguistic representations},
  author={Su, Weijie and Zhu, Xizhou and Cao, Yue and Li, Bin and Lu, Lewei and Wei, Furu and Dai, Jifeng},
  journal={arXiv preprint arXiv:1908.08530},
  year={2019}
}

@inproceedings{jain2021perturb,
  title={Perturb, Predict \& Paraphrase: Semi-Supervised Learning using Noisy Student for Image Captioning.},
  author={Jain, Arjit and Samala, Pranay Reddy and Jyothi, Preethi and Mittal, Deepak and Singh, Maneesh Kumar},
  booktitle={IJCAI},
  pages={758--764},
  year={2021}
}

@article{kuang2023dlip,
  title={DLIP: Distilling Language-Image Pre-training},
  author={Kuang, Huafeng and Wu, Jie and Zheng, Xiawu and Li, Ming and Xiao, Xuefeng and Wang, Rui and Zheng, Min and Ji, Rongrong},
  journal={arXiv preprint arXiv:2308.12956},
  year={2023}
}

@inproceedings{voxelnet,
  title={Voxelnet: End-to-end learning for point cloud based 3d object detection},
  author={Zhou, Yin and Tuzel, Oncel},
  booktitle={Proceedings of the IEEE conference on computer vision and pattern recognition},
  pages={4490--4499},
  year={2018}
}

@article{second,
  title={Second: Sparsely embedded convolutional detection},
  author={Yan, Yan and Mao, Yuxing and Li, Bo},
  journal={Sensors},
  volume={18},
  number={10},
  pages={3337},
  year={2018},
  publisher={MDPI}
}

@inproceedings{octnet,
  title={Octnet: Learning deep 3d representations at high resolutions},
  author={Riegler, Gernot and Osman Ulusoy, Ali and Geiger, Andreas},
  booktitle={Proceedings of the IEEE conference on computer vision and pattern recognition},
  pages={3577--3586},
  year={2017}
}

@inproceedings{pointnet,
  title={Pointnet: Deep learning on point sets for 3d classification and segmentation},
  author={Qi, Charles R and Su, Hao and Mo, Kaichun and Guibas, Leonidas J},
  booktitle={Proceedings of the IEEE conference on computer vision and pattern recognition},
  pages={652--660},
  year={2017}
}

@article{pointnet++,
  title={Pointnet++: Deep hierarchical feature learning on point sets in a metric space},
  author={Qi, Charles Ruizhongtai and Yi, Li and Su, Hao and Guibas, Leonidas J},
  journal={Advances in neural information processing systems},
  volume={30},
  year={2017}
}

@inproceedings{votenet,
  title={Votenet: A deep learning label fusion method for multi-atlas segmentation},
  author={Ding, Zhipeng and Han, Xu and Niethammer, Marc},
  booktitle={Medical Image Computing and Computer Assisted Intervention--MICCAI 2019: 22nd International Conference, Shenzhen, China, October 13--17, 2019, Proceedings, Part III 22},
  pages={202--210},
  year={2019},
  organization={Springer}
}

@misc{contextual,
      title={Contextual Modeling for 3D Dense Captioning on Point Clouds}, 
      author={Yufeng Zhong and Long Xu and Jiebo Luo and Lin Ma},
      year={2022},
      eprint={2210.03925},
      archivePrefix={arXiv},
      primaryClass={cs.CV}
}

@inproceedings{brnet,
  title={Back-tracing representative points for voting-based 3d object detection in point clouds},
  author={Cheng, Bowen and Sheng, Lu and Shi, Shaoshuai and Yang, Ming and Xu, Dong},
  booktitle={Proceedings of the IEEE/CVF Conference on Computer Vision and Pattern Recognition},
  pages={8963--8972},
  year={2021}
}

@inproceedings{h3dnet,
  title={H3dnet: 3d object detection using hybrid geometric primitives},
  author={Zhang, Zaiwei and Sun, Bo and Yang, Haitao and Huang, Qixing},
  booktitle={Computer Vision--ECCV 2020: 16th European Conference, Glasgow, UK, August 23--28, 2020, Proceedings, Part XII 16},
  pages={311--329},
  year={2020},
  organization={Springer}
}

@inproceedings{rbgnet,
  title={Rbgnet: Ray-based grouping for 3d object detection},
  author={Wang, Haiyang and Shi, Shaoshuai and Yang, Ze and Fang, Rongyao and Qian, Qi and Li, Hongsheng and Schiele, Bernt and Wang, Liwei},
  booktitle={Proceedings of the IEEE/CVF Conference on Computer Vision and Pattern Recognition},
  pages={1110--1119},
  year={2022}
}

@inproceedings{3detr,
  title={An end-to-end transformer model for 3d object detection},
  author={Misra, Ishan and Girdhar, Rohit and Joulin, Armand},
  booktitle={Proceedings of the IEEE/CVF International Conference on Computer Vision},
  pages={2906--2917},
  year={2021}
}

@misc{qi2019deep,
      title={Deep Hough Voting for 3D Object Detection in Point Clouds}, 
      author={Charles R. Qi and Or Litany and Kaiming He and Leonidas J. Guibas},
      year={2019},
      eprint={1904.09664},
      archivePrefix={arXiv},
      primaryClass={cs.CV}
}

@inproceedings{fcaf3d,
  title={Fcaf3d: Fully convolutional anchor-free 3d object detection},
  author={Rukhovich, Danila and Vorontsova, Anna and Konushin, Anton},
  booktitle={European Conference on Computer Vision},
  pages={477--493},
  year={2022},
  organization={Springer}
}

@article{cagroup3d,
  title={Cagroup3d: Class-aware grouping for 3d object detection on point clouds},
  author={Wang, Haiyang and Dong, Shaocong and Shi, Shaoshuai and Li, Aoxue and Li, Jianan and Li, Zhenguo and Wang, Liwei and others},
  journal={Advances in Neural Information Processing Systems},
  volume={35},
  pages={29975--29988},
  year={2022}
}

@misc{ptv2,
      title={Point Transformer V2: Grouped Vector Attention and Partition-based Pooling}, 
      author={Xiaoyang Wu and Yixing Lao and Li Jiang and Xihui Liu and Hengshuang Zhao},
      year={2022},
      eprint={2210.05666},
      archivePrefix={arXiv},
      primaryClass={cs.CV}
}

@misc{scst,
      title={Self-critical Sequence Training for Image Captioning}, 
      author={Steven J. Rennie and Etienne Marcheret and Youssef Mroueh and Jarret Ross and Vaibhava Goel},
      year={2017},
      eprint={1612.00563},
      archivePrefix={arXiv},
      primaryClass={cs.LG}
}

@misc{minkunet,
      title={4D Spatio-Temporal ConvNets: Minkowski Convolutional Neural Networks}, 
      author={Christopher Choy and JunYoung Gwak and Silvio Savarese},
      year={2019},
      eprint={1904.08755},
      archivePrefix={arXiv},
      primaryClass={cs.CV}
}

@misc{swin3d,
      title={Swin3D: A Pretrained Transformer Backbone for 3D Indoor Scene Understanding}, 
      author={Yu-Qi Yang and Yu-Xiao Guo and Jian-Yu Xiong and Yang Liu and Hao Pan and Peng-Shuai Wang and Xin Tong and Baining Guo},
      year={2023},
      eprint={2304.06906},
      archivePrefix={arXiv},
      primaryClass={cs.CV}
}
}

%
\definecolor{cvprblue}{rgb}{0.21,0.49,0.74}

\appendix


\section{Additional Ablations}
\label{sec:supply_ablation}
\subsection{Ablation Study on Backbone Capacity}
In our VoxelCapNet, we find that the backbone capacity is an important factor 
in determining the final performance. In Table \ref{exp:channel}, we take 
SparseUNet\cite{minkunet} as our VoxelBackbone and use different channels 
for each layer of SparseUNet. SparseUNet consists of 4 downsampling and 4 upsampling layers.
Experiment (2) uses the default channel configuration of SparseUNet, \textit{i.e.}, [32, 64, 128, 256, 256, 128, 96].
When we increase the backbone channels are increased to [48, 96, 192, 384, 384, 192, 96], 
the caption performance is improved from 65.17\% C@0.5IoU to 66.35\% C@0.5IoU.
Besides, with backbone channels decreasing, the performance drops.
The results showcase that a large backbone capacity can lead to a better performance.

\subsection{Ablation Study on Loss Weight}
Because our approach trains the detection task and the caption task jointly, 
it is necessary to select proper loss weights to balance the two tasks.
In Table \ref{exp:weight}, we fix the loss weight of caption loss to 1 and ablate the 
loss weight of detection loss. The results suggest that our method achieves the 
best performance when using a detection loss weight of 1.0. 

\subsection{NMS Threshold of the Detection Head}
Non-maximum suppression (NMS) is used in the detection head of our VoxelCapNet to drop duplicate predictions.
Specifically, if a high-confidence predicted box has an IoU larger than a pre-defined threshold with another predicted box, 
it will suppress and drop the box.
In Table \ref{exp:nms_thres}, we provide an ablation study on the IoU threshold of NMS.
A small IoU threshold of NMS can lead to a large number of false negatives, while 
a large IoU threshold of NMS can decrease the recall.
The results show that VoxelCapNet with a threshold of 0.5 can perform best, achieving 69.59\% C@0.5IoU.


\begin{table}[!t]   
    \caption{Ablation study on backbone capacity.}
    \vspace{-2mm}
    \centering
    \resizebox{1.05\linewidth}{!}{
    \renewcommand\arraystretch{1.2}
        \begin{tabular}{c|c|cccc}
        \toprule
        ID  & Backbone Channels  & C@0.5   & B-4@0.5   & M@0.5   & R@0.5 \\ \hline
        (1) & 24, 48,  96, 192, 192,  96, 96 & 64.62   & 38.68     & 27.83   & 59.94 \\
        (2) & 32, 64, 128, 256, 256, 128, 96 & 65.17   & 38.57     & 27.75   & 59.83 \\
        (3) & 48, 96, 192, 384, 384, 192, 96 & 66.35   & 39.35     & 28.01   & 60.32 \\
        \bottomrule 
        \end{tabular}
    }
    \label{exp:channel}
    \vspace{-2mm}
\end{table}

\begin{table}[!t]   
    \caption{Ablation study on detection loss weight $\alpha$.}\label{exp:weight}
    \vspace{-2mm}
    \centering
    \resizebox{.88\linewidth}{!}{
        \begin{tabular}{c|c|cccc}
        \toprule
        ID  & $\alpha$  & C@0.5   & B-4@0.5   & M@0.5   & R@0.5 \\ \hline
        (1) & 0.5      & 68.64   & 39.06     & 27.90   & 59.74 \\  
        (2) & 1        & 69.59   & 39.24     & 28.05   & 60.04 \\
        (3) & 2        & 68.61   & 39.70     & 28.03   & 59.90 \\
        \bottomrule
        \end{tabular}
    }
    \vspace{-2mm}
\end{table}

\begin{table}[!t]   
    \caption{Ablation study on NMS threshold $\tau$ of detection head.}\label{exp:nms_thres}
    \vspace{-2mm}
    \centering
    \resizebox{.88\linewidth}{!}{
        \begin{tabular}{c|c|cccc}
        \toprule
        ID  & $\tau$  & C@0.5   & B-4@0.5   & M@0.5   & R@0.5 \\ \hline
        (1) & 0.3      & 65.46   & 37.53     & 27.45   & 58.26 \\
        (2) & 0.5        & 69.59   & 39.24     & 28.05   & 60.04 \\
        (3) & 0.7       & 64.83   & 40.40     & 28.36   & 60.88 \\ 
        \bottomrule 
        \end{tabular}
    }
    \vspace{-2mm}
\end{table}


\begin{figure*}[t]
	\begin{center}
		\setlength{\fboxrule}{0pt}
		\fbox{\includegraphics[width=.96\textwidth]{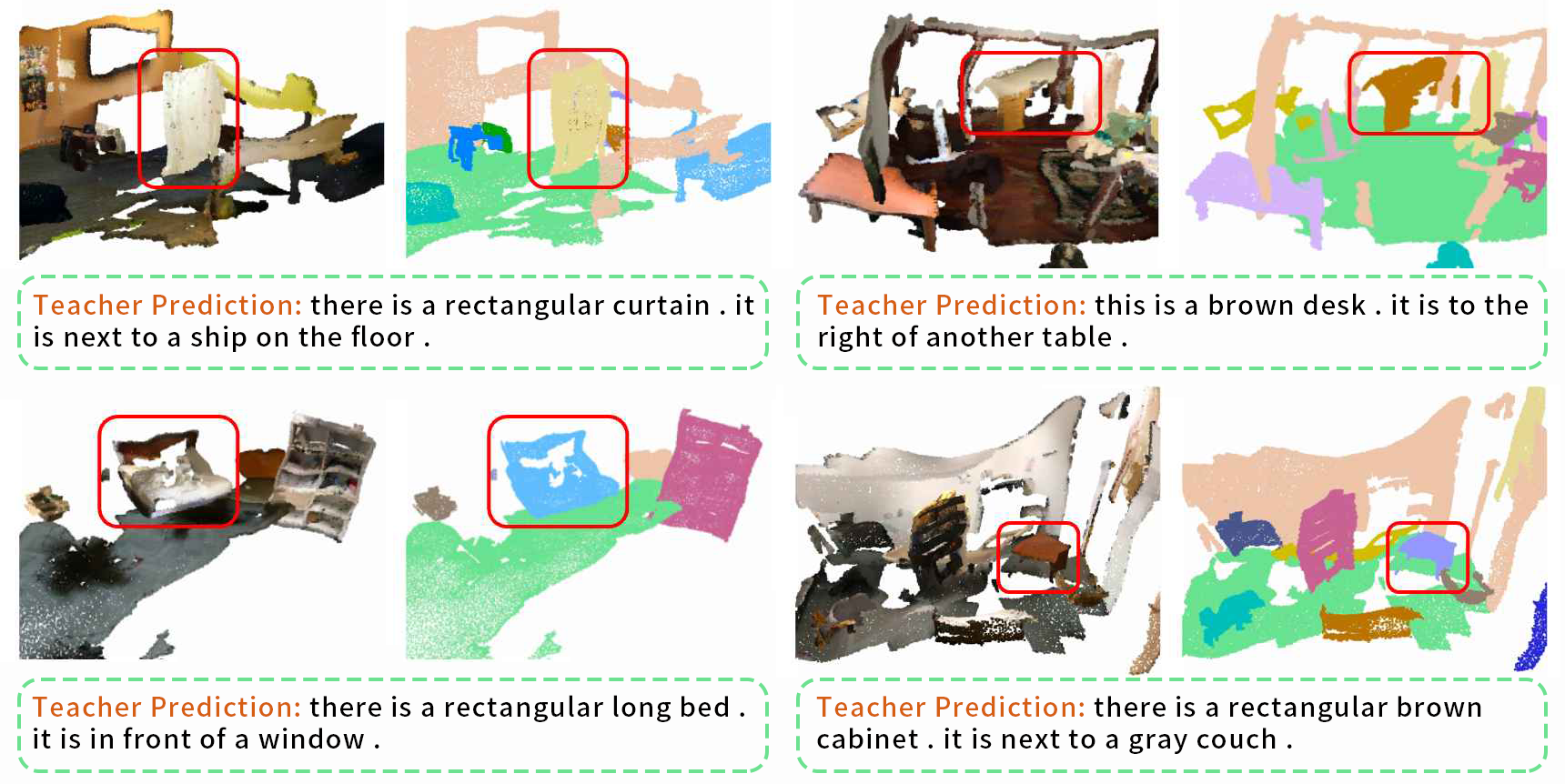}}
	\end{center}
	\vspace{-6mm}
	\caption{Visualization of pseudo frames generated by the proposed PseudoCap.}
	\label{fig:pseudo_frame}
	\vspace{-2mm}
\end{figure*}

\section{Additional Details}
\label{sec:supply_details}
In this section, we elaborate on Instance Mixing, the process of generating pseudo frames, in detail.
For the real frames that are selected for PseudoCap, we first keep their background point clouds (wall and floor) and filter their foreground point clouds. 
During PseudoCap, we pre-define a number for each category (1 by default) to control the number of instances we select.
For the selected instance, we randomly set a location within the coordinate range of the current scene as the center to paste the instance.
In particular, instances from specific categories, \textit{i.e.}, table and chair, should be placed on the floor following the real scenes.
In order to prevent pasted instances from overlapping with existing instances in the scene, we need to perform collision detection with a pre-defined IoU threshold.
Besides, with more instances joining, the scene will become crowded. 
To prevent instances of categories at the end of the queue from being unable to be selected, the category queue will be shuffled every time we perform Instance Mixing. 

\section{Additional Qualitative Results}
\label{sec:supply_visualization}

\begin{figure*}[h]
	\begin{center}
		\setlength{\fboxrule}{0pt}
		\fbox{\includegraphics[width=1.\textwidth]{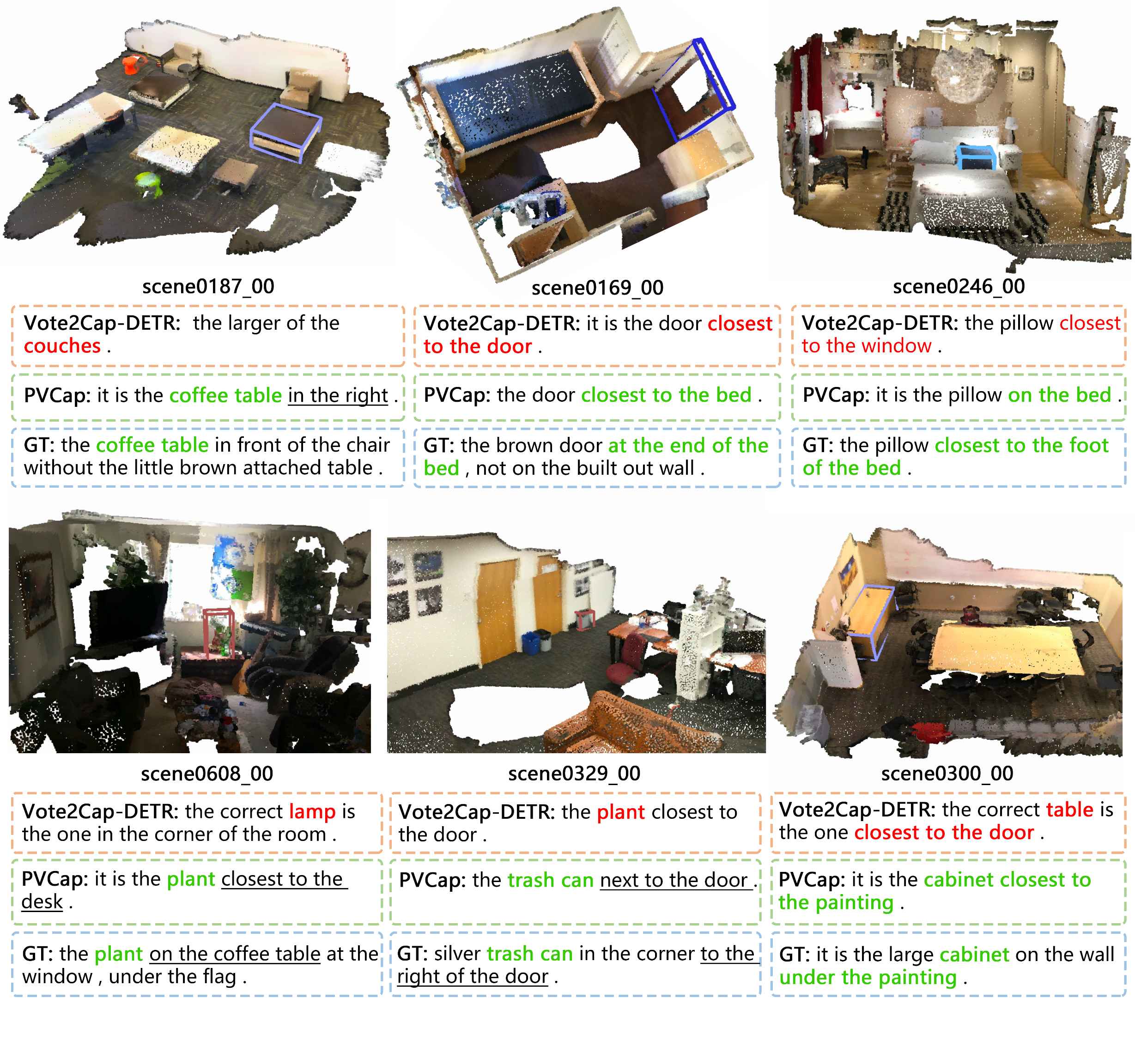}}
	\end{center}
	\vspace{-10mm}
	\caption{Qualitative Comparison on Nr3D Dataset}
	\label{fig:nr3d_vis}
	\vspace{-2mm}
\end{figure*}

\subsection{Visualization for Pseudo Frames}
In our PseudoCap, we crop instances of real frames in the dataset to build an instance database. 
Then, we randomly select some instances from the database and mix them with empty real frames 
to generate pseudo frames with diverse spatial layouts.
As shown in Figure \ref{fig:pseudo_frame}, we provide visualization of these pseudo frames 
and their pseudo labels predicted by the teacher caption model. 
From the visualization, we can find that different instances scatter in the pseudo frames and 
the spatial layouts are various. 
Instances in pseudo frames can cross over the wall, and their orientations are arbitrary.
These make pseudo frames differ from the real scene, but they can provide sufficient supervision about 
spatial relations and play a role in regularization to improve the caption model.

\subsection{Qualitative Comparison on Nr3D Dataset}
To qualitatively evaluate the caption result, we compare our method with Vote2Cap-DETR 
\cite{vote2cap} and ground truth on Nr3D, as demonstrated in Figure \ref{fig:nr3d_vis}. 
The wrong descriptions are marked in red and the corresponding correct descriptions are in green. 
We also underline the surrounding information to showcase the ability of our method on environment description.
From the visualization results, we can find that our method can provide a more accurate classification of 
foreground objects than Vote2Cap-DETR, which can be attributed to the strong backbone and detection head 
in VoxelCapNet. Moreover, our method also exhibits stronger environment description ability, which 
can be attributed to the diverse training data provided by PseudoCap. 


%

\end{document}